\documentclass[11pt,a4paper]{article}
\usepackage[hyperref]{acl2019}
\usepackage{times}
\usepackage{latexsym}

\usepackage{graphicx}
\usepackage{float}

\usepackage{url}

\aclfinalcopy % Uncomment this line for the final submission
\def\aclpaperid{716} %  Enter the acl Paper ID here

\title{CNNs found to jump around more skillfully than RNNs:\\Compositional generalization in seq2seq convolutional networks}

\author{Roberto Dess\`i \\
  CIMeC, University of Trento \\
  \texttt{roberto.dessi@studenti.unitn.it} \\\And
  Marco Baroni \\
  ICREA \\
  Facebook AI Research \\
  \texttt{mbaroni@fb.com} \\}

\date{}

\begin{document}

\maketitle

\begin{abstract}
  \citet{Lake:Baroni:2017} introduced the SCAN dataset probing the
  ability of seq2seq models to capture compositional generalizations,
  such as inferring the meaning of \emph{``jump around''} 0-shot from
  the component words. Recurrent networks (RNNs) were found to
  completely fail the most challenging generalization cases. We test
  here a convolutional network (CNN) on these tasks, reporting hugely improved performance with respect to
  RNNs. Despite the big improvement, the
  CNN has however not induced systematic rules, suggesting that the difference
  between compositional and non-compositional behaviour is not
  clear-cut.
\end{abstract}

\section{Introduction}
\label{sec:intro}

Recent deep neural network successes rekindled classic
debates on their natural language processing abilities
\cite[e.g.,][]{Kirov:Cotterell:2018,McCoy:etal:2018,Pater:2018}. \citet{Lake:Baroni:2017}
and \citet{Loula:etal:2018} proposed the SCAN challenge to
directly assess the ability of sequence-to-sequence networks to
perform systematic, compositional generalization of linguistic
rules. Their results, and those of \citet{Bastings:etal:2018}, have
shown that modern recurrent networks (gated RNNs, such as LSTMs and GRUs) generalize well
to new sequences that resemble those encountered in training,
but achieve very low performance when generalization must be
supported by a systematic compositional rule, such as ``to X twice 
you X and X'' (e.g., to jump twice, you jump and jump again).

Non-recurrent models, such as convolutional neural networks
\cite[CNNs,][]{kalchbrenner:etal:2016, gehring:etal:2016,
  gehring:etal:2017} and self-attentive models
\cite{vaswani:etal:2017, chen:etal:2018} have recently reached
comparable or better performance than RNNs on machine translation and
other benchmarks. Their linguistic properties are however still
generally poorly understood. \newcite{Tang:etal:2018} have shown that
RNNs and self-attentive models are better than CNNs at capturing
long-distance agreement, while self-attentive networks excel at word
sense disambiguation. In an extensive comparison,
\newcite{Bai:etal:2018} showed that CNNs generally outperform RNNs,
although the differences were typically not huge. We evaluate here an out-of-the-box CNN on the most
challenging SCAN tasks, and we uncover the surprising fact that
\emph{CNNs are dramatically better than RNNs at compositional
  generalization}. As they are more cumbersome to train, we leave
testing of self-attentive networks to future work.

\section{SCAN}
\label{sec:setup}

SCAN studies compositionality in a simple command
execution environment framed as a supervised sequence-to-sequence
task. The neural network receives word sequences as input, and has to
produce the correspondence action sequence. Examples are
given in Table \ref{table:examples}.
  \newcite{Lake:Baroni:2017} originally introduced 4
train/test splits, of which we consider 2.\footnote{We also
  tested our CNNs on SCAN's \emph{length} split, where test commands
  require systematically longer actions than the training
  ones. Accuracy was near 0\%, as the learned positional
  embeddings of our CNN architecture do not 
  generalize beyond  training lengths. We leave the
  investigation of more flexible positional encodings \cite[as in,
  e.g.,][]{vaswani:etal:2017} to future work. We also experimented
  with SCAN's \emph{turn left} split, obtaining near-perfect
  generalization. As RNNs were already performing very well in this split,
  we focus in the paper on the more challenging
  \emph{jump} case.} In the \emph{random} split, the training set
includes 80\% of randomly selected distinct SCAN commands, with the
remaining 20\% in the test set. This requires generalization,
as no test command is encountered in training, but there is no
systematic difference between the commands in the two sets.  In the
\emph{jump} split, the \emph{jump} command is only seen in isolation
during training, and the test set consists of all composite commands
with \emph{jump}. A system able to extract compositional rules (such
as ``\emph{X twice} means to X and X'') should have no problem
generalizing them to a new verb, as in this
split. \newcite{Loula:etal:2018} proposed a set of new SCAN splits,
the most challenging one being the \emph{around-right}
split.  The training partition contains examples of \emph{around} and
\emph{right}, but never in combination. The test set
contains all possible \emph{around right} commands. Loula and
colleagues want to test ``second-order modification'', as models
must learn how to compositionally apply the \emph{around}
function to \emph{right}, which is in turn a first-order function
modifying simple action verbs.

\begin{table}[t!]
    \footnotesize
    \begin{center}
%      \begin{tabular}{| l | l | l |}
      \begin{tabular}{| c | p{2.2cm} | p{2.2cm} |}
            \hline \textbf{Split} & \textbf{Train Command} & \textbf{Test Command} \\ \hline
            \textit{random} & \textit{walk opposite left}; \textit{turn left twice and look} & 
                \textit{walk and jump right twice}; \textit{run and run thrice}  \\
            \hline
            \textit{jump} & \textit{\underline{jump}}; \textit{turn left twice \underline{after look}}  & 
            \textit{turn left twice \underline{after jump}}; \textit{run twice \underline{and jump}} \\% jump and turn opposite right} \\
            \hline
            \textit{around-right} & \textit{jump \underline{around left}}; \textit{turn \underline{opposite right} twice} & \textit{walk \underline{around right}};
            \textit{look \underline{around right} and jump left} \\
            %\textit{around-right} & \textit{jump around left thrice}; \textit{turn opposite right twice} & \textit{walk around right};
            %\textit{look around right and jump left} \\
            \hline
        \end{tabular} 
    \end{center}
    \caption{\label{table:examples} Training and test examples for the three splits used in our experiments.}
\end{table}

\section{Experimental setup}
\paragraph{Model} We use the fully convolutional encoder-decoder model
of \newcite{gehring:etal:2017} out of the box, using version 0.6.0 of the fairseq 
toolkit.\footnote{\url{https://github.com/pytorch/fairseq}} The model
uses convolutional filters and Gated Linear Units
\cite{dauphin:etal:2016} along with an attention mechanism
that connects the encoder and the decoder.  Attention is computed
separately for each encoder layer, and produces weighted sums over
encoder input embeddings and encoder outputs. See the original paper
for details.
\paragraph{Training} The shift in distribution between training and
test splits makes SCAN unsuitable for validation-set tuning. Instead,
following \newcite{Lake:Baroni:2017} and \newcite{Loula:etal:2018}, we
train on 100k random samples with replacement from the
training command set. We explore different batch sizes (in terms of number of tokens per 
batch: 25, 50, 100, 200, 500, 1000), learning rates (0.1, 0.01, 0.001), layer dimensionalities
(128, 256, 512), layer numbers (6 to 10), convolutional
kernel width (3, 4, 5) and amount of dropout used (0, 0.25, 0.5).
For all other hyperparameters, we accept recommended/default fairseq values. Each configuration is run with 5
seeds, and we report means and standard deviations.

\section{Results}
\label{sec:experiments}

\begin{table}[tb]
  \begin{footnotesize}
    \begin{center}
%        \scalebox{0.96}{
          % \begin{tabular}{l | p{2cm} | p{2cm} | p{2cm}}
          \begin{tabular}{l | r | r | r}
             & \textbf{random} & \textbf{jump} & \textbf{around-right} \\ \hline
             LSTM & 99.8 & 1.2 & 2.5$\pm$2.7 \\
             GRU & \textbf{100.0}$\pm$0.0 & 12.5$\pm$6.6 & --  \\
             \hline
              CNN & \textbf{100.0}$\pm$0.0 & \textbf{69.2}$\pm$8.2 & \textbf{56.7}$\pm$10.2 \\
          \end{tabular} 
%    }
    \end{center}
  \end{footnotesize}
  \caption{Test accuracy (\%) on SCAN splits (means across 5 seeds,
    with standard deviation if available). Top LSTM results from
    \newcite{Lake:Baroni:2017}/\newcite{Loula:etal:2018}, GRU from
      \newcite{Bastings:etal:2018}.}
\label{table:main_results} 
\end{table}

Our main results are in Table \ref{table:main_results}. CNNs, like
RNNs, succeed in the \emph{random} split, and achieve much higher accuracy (albeit still far from being perfect)  in the challenging \emph{jump} and \emph{around-right}
splits.

\begin{figure}[tb]
    \centering
    \includegraphics[width=.45\textwidth,keepaspectratio]{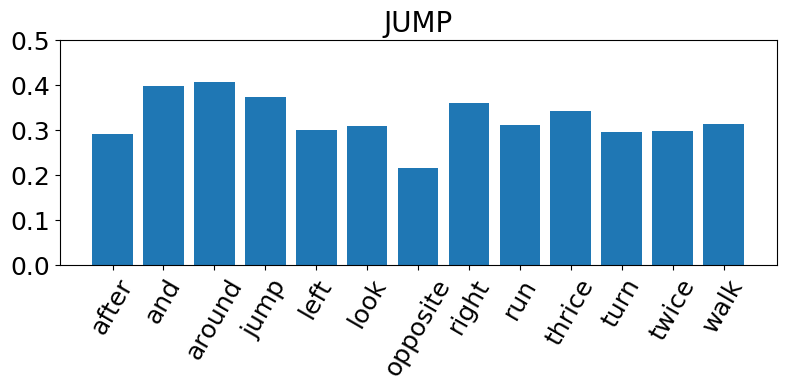}
    \includegraphics[width=.45\textwidth,keepaspectratio]{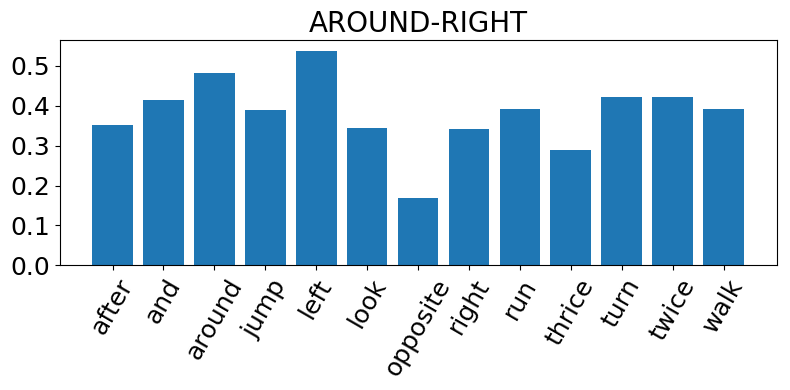}
    \caption{Proportion of commands with a certain command word (over total commands with that word) wrongly executed by best CNNs.}
    \label{fig:error_distributions}
\end{figure}

\begin{figure}[tb]
    \includegraphics[width=.45\textwidth,keepaspectratio]{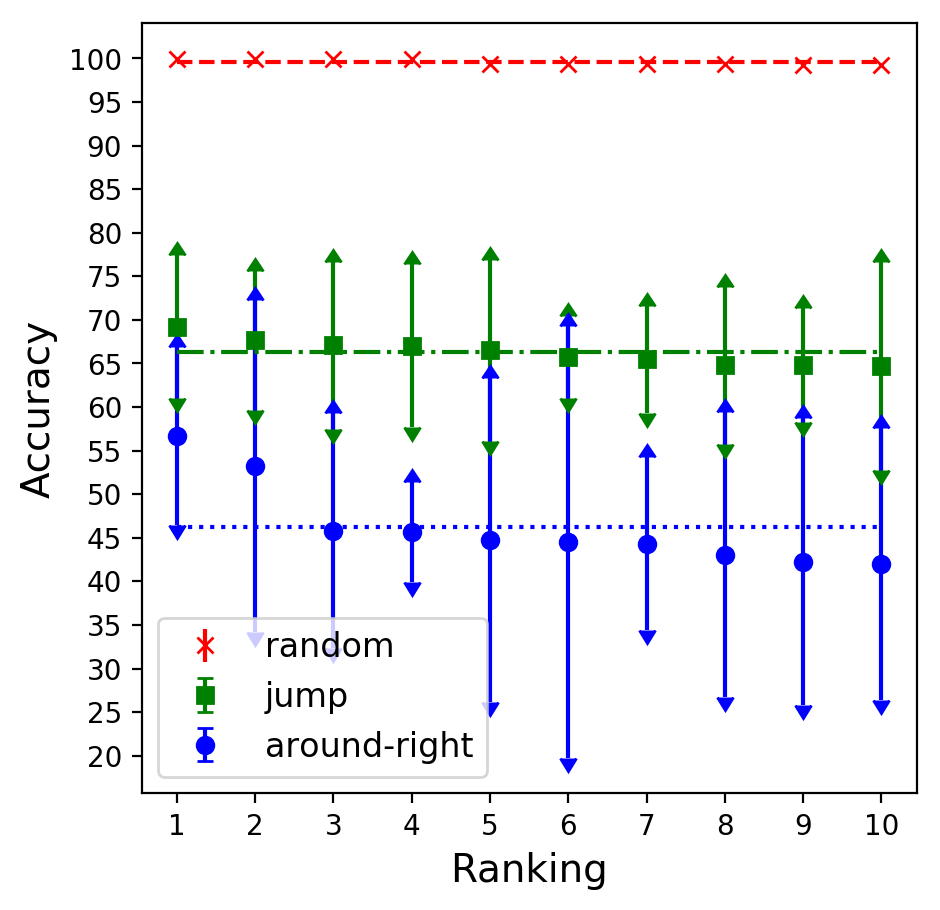}
    \centering
    \caption{Accuracies (\%) of top-10 models on \emph{random}, \emph{jump} and \emph{around-right}. Arrows
      denote standard deviations, dashed lines average accuracy across top-10.
      % \mb{means across what? Also, it would be better if
      %legend order was random/jump/around-right}
      }
    \label{fig:exp1}
\end{figure}

The SCAN tasks should be easy for a system that learned the right
composition rules. Perhaps, CNNs do not achieve 100\% accuracy because
they only learned a subset of the necessary rules. For example, they
might correctly interpret the new expression \emph{jump twice} because
they induced a \emph{X twice} rule at training time, but fail
\emph{jump thrice} because they missed the corresponding \emph{X
  thrice} rule. Since SCAN semantic composition rules are associated
with single words in input commands, we can check this hypothesis by
looking at error distribution across input words. It turns
out (Fig.~\ref{fig:error_distributions}) that errors are not
associated to specific input commands. Error proportion is instead relatively
stable across command words. Direct inspection reveals no traces of
systematicity: errors cut across composition rules. Indeed, we
often find minimal pairs in which changing one action verb with
another (distributionally equivalent in SCAN) turns a correctly
executed command into a failed one. For example, in the \emph{jump}
split, the CNN correctly executes \emph{jump left after walk}, but
fails \emph{jump left after run} (jumping is
forgotten). Analogously, in the \emph{around-right} split, \emph{run
  around right} is correctly executed, but ``\emph{walk around
  right}'' is not (the network stops too early).
\paragraph{Robustness} Fig.~\ref{fig:exp1} shows a big difference in
stability between \emph{random} and the other splits across top
hyperparameter configurations. The \emph{random} results are very
stable. \emph{Jump} accuracy is relatively stable across
hyperparameters, but has large variance across initialization
seeds. For the most challenging \emph{around-right} split, we observe
instability both across seeds and hyperparameters (although even the
lowest end of the reported accuracies is well above best RNN
performance in the corresponding experiments). Another question is
whether the best configurations are shared, or each split requires an
\emph{ad-hoc} hyperparameter choice. We find that there are configurations
that achieve good performance across the splits. In particular, the
\emph{best overall configuration}, found by minimizing ranks across
splits, has 0.01 learning rate, 25-tokens batch size, 0.25
dropout, 6 layers, 512 layer dimensionality, and kernels of width
5. Such model was 13th best (of about 2.5K explored) on the
\emph{random} split (with mean cross-seed accuracy of 99.92\%, off
by 0.05\% from top configuration), 32th on the \emph{jump} split
(60.67\% mean accuracy, off by 8.62\%), and 2nd in the
\emph{around-right} split (mean 53.25\% accuracy, off by 3.45\%).

\paragraph{Kernel width}
\label{subsec:exp2}

One important difference between recurrent and convolutional
architectures is that CNN kernel
width imposes a strong prior on the window of elements to be
processed together. We conjecture that relatively wide encoder and
decoder widths, by pushing the network to keep wider
contexts into account, might favour the acquisition of template-based
generalizations, and hence better compositionality. To investigate
this, we varied encoder and decoder widths of the best-overall model
between 1 and 5.\footnote{At least on the encoder side, larger widths
  seem excessive, as the longest commands are 9-word-long.}

\begin{figure*}[tb]
    \centering
    \includegraphics[width=\textwidth,keepaspectratio]{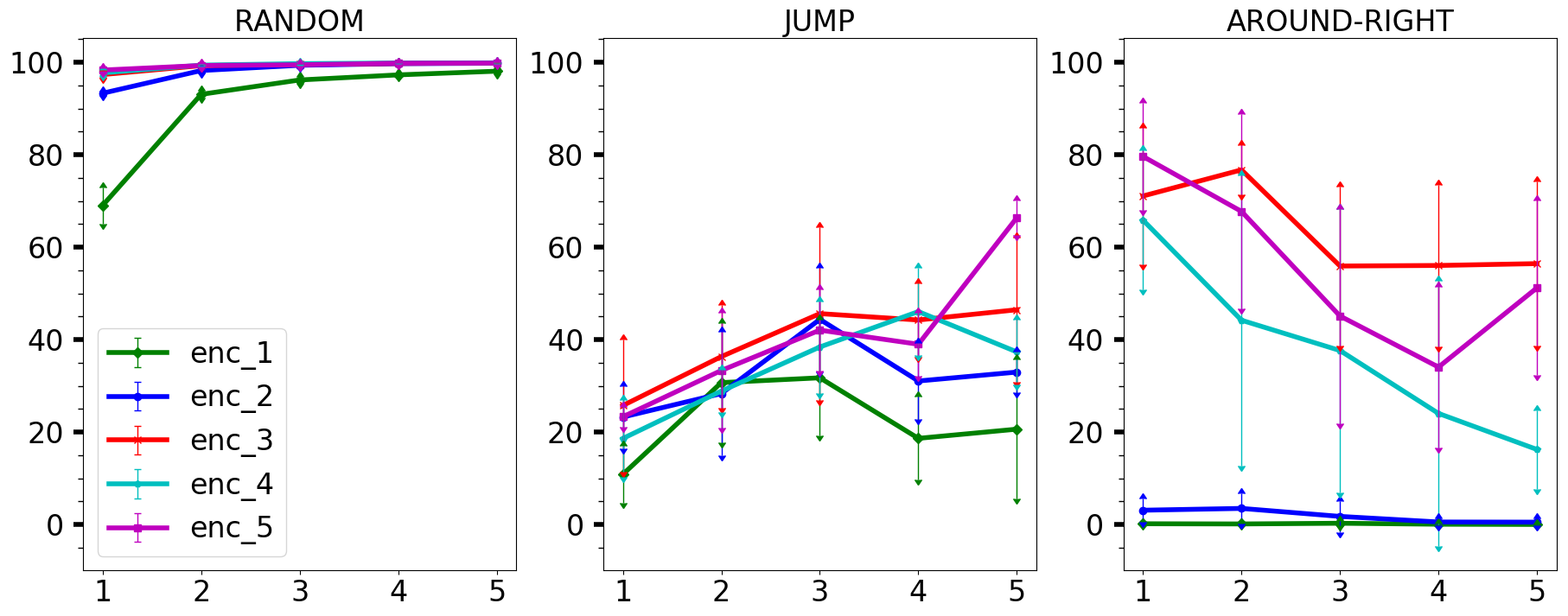}
    \caption{Mean accuracies (\%) across 5 seeds, in function of decoder (x axis) and encoder (colors) kernel widths. Arrows denote standard deviations. Best viewed in color.
    %More natural order would be random/jump/around-right.
%    \mb{Also crop actual file and enlarge figure.}
    }
    \label{fig:kernel_exp}
\end{figure*}

Fig.~\ref{fig:kernel_exp} shows that the \emph{random} split confirms
our expectations, as both wider encoder and decoder windows improve
performance. The \emph{jump} results follow the same trend, although
in a less clear-cut way. Still, the narrowest encoder-decoder
combination has the worst performance, and the widest one the top
one. For the \emph{around-right} split, it is also better to use the
widest encoder, but top performance is achieved with the
\emph{narrowest} decoder (width$=$1). Indeed, with the narrow decoder
we obtain \emph{around-right} accuracies that are even above the
absolute-best \emph{jump}-split performance. Since the novel output
templates in the \emph{around-right} split are by construction long
(they involve executing an \emph{around} command that requires
repeating an action 4 times), we would have rather expected models
keeping track of a larger decoding window to fare better, particularly
in this case. We tried to gain some insight on the attested behaviour
by looking at performance distribution in function of input and output
length, failing to detect different patterns in the wide-decoder
\emph{jump} model vs.~the narrow-decoder \emph{around-right} model
(analysis not reported here for space reasons). Looking qualitatively
at the errors, we note that, for both splits, the narrower decoder
tends to skip trajectory sub-chunks (e.g., executing ``\emph{jump
  around right}'' with 3 instead of 4 right turns followed by jumps),
whereas the wider kernel is more likely to substitute actions (e.g.,
turning left instead of right) than undershooting the length. This
impressionistic observation is supported by the fact that, for both
splits, the narrow-kernel errors have considerably larger variance
than the wide-kernel errors with respect to ground-truth length,
indicating that, with narrow decoder kernel, the model is less stable in terms of output sequence length. This, however, only confirms our
 conjecture that a wider decoder kernel helps length management.
We still have no insight on why the narrower kernel should be better on
the \emph{around-right split}.
\paragraph{Multi-layer attention}
\label{subsec:exp3}

\begin{figure}[tb]
    \centering
    \includegraphics[width=.5\textwidth,keepaspectratio]{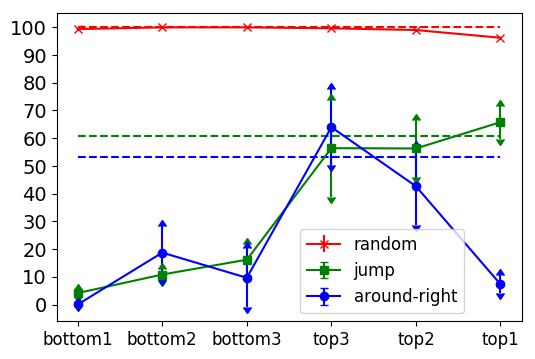}
    \caption{Accuracy (\%) of overall-best model with attention
      only from first layer (\emph{bottom1}), first two layers
      (\emph{bottom2}), \ldots, last two layers (\emph{top2}), top
      layer only (\emph{top1}). Means and standard deviations across 5
      seeds. Dashed lines show full multi-layer attention results.}
    \label{fig:exp3}
\end{figure}

The fairseq CNN has attention from all layers of the
decoder. Is the possibility to focus on different aspects of the input
while decoding from different layers crucial to its better
generalization skills? Fig.~\ref{fig:exp3} reports
accuracies when applying attention from a subset of the 6
layers only. The \emph{random} split differences are minimal,
but ablating attentions greatly affects performance on the
compositional splits (although, in both cases, there is a single ablated
configuration that is as good as the full setup).

\section{Conclusion}

Compared to the RNNs previously tested in the
literature, the out-of-the-box fairseq CNN architecture reaches
dramatically better performance on the SCAN compositional
generalization tasks. The CNN
is however not  learning rule-like compositional generalizations, as its
mistakes are non-systematic and they are evenly spread across different
commands. Thus, the CNN  achieved a considerable
degree of generalization, even on an explicitly compositional
benchmark, without something akin to rule-based reasoning. Fully
understanding generalization of deep seq2seq models might require a less
clear-cut view of the divide between statistical pattern matching and
symbolic composition. In future work, we would like to further our insights on the
CNN aspects that are crucial for the task,  our preliminary
analyses of kernel width and attention.
\\
Concerning the comparison with RNNs, the best LSTM architecture of
Lake and Baroni has two 200-dimensional layers, and it is consequently
more parsimonious than our best CNN (1/4 of parameters). In informal
experiments, we found shallow CNNs incapable to handle even the
simplest \emph{random} split. On the other hand, it is hard to train
very deep LSTMs, and it is not clear that the latter models need the
same depth CNNs require to ``view'' long sequences. We
leave a proper formulation of a tighter comparison to future work.

\section*{Acknowledgements}
We thank Brenden Lake, Michael Auli, Myle Ott, Jo\~ao Loula, Joost Bastings and the reviewers for comments and advice.

\bibliography{marco,roberto}
\bibliographystyle{natbib}

\end{document}